\documentclass[runningheads]{llncs}
\usepackage{graphicx}
\usepackage{amsmath,amssymb} 
\usepackage{color}
\usepackage{soul}
\usepackage{multirow}
\usepackage{graphicx}
\usepackage{booktabs}
\usepackage{svg}
\usepackage{caption}
\usepackage[caption=false]{subfig}

\DeclareMathOperator*{\avg}{\mathrm{average}}
\DeclareMathOperator*{\dca}{\mathrm{dca}}

\newcommand{\mcrot}[4]{\multicolumn{#1}{#2}{\rlap{\rotatebox{#3}{#4}~}}} 

\newcommand*{\twoelementtable}[3][l]%
{%
	\begin{tabular}[t]{@{}#1@{}}%
		#2\tabularnewline
		#3%
	\end{tabular}%
}

\begin{document}
\pagestyle{headings}
\mainmatter
	
\def\ACCV20SubNumber{715}  
	
\title{OpenTraj: Assessing Prediction Complexity in Human Trajectories Datasets}
\titlerunning{OpenTraj: Assessing Prediction Complexity in HTP Datasets}
	
\author{Javad Amirian\inst{1} \and
		Bingqing Zhang\inst{2} \and
		Francisco Valente Castro\inst{3}\and
		Juan Jos\'e Baldelomar\inst{3}\and
		Jean-Bernard Hayet\inst{3}\and
		Julien Pettr{\'e}\inst{1}
		}
\authorrunning{J. Amirian et al.}

\institute{Univ Rennes, Inria, CNRS, IRISA, France \and
		   University College London, UK \and
		   CIMAT, A.C., M{\'e}xico }		
	
\maketitle

\begin{abstract}
Human Trajectory Prediction (HTP) has gained much momentum in the last years and many solutions have been proposed to solve it. Proper benchmarking being a key issue for comparing methods, this paper addresses the question of evaluating how complex is a given dataset with respect to the prediction problem. For assessing a dataset complexity, we define a series of indicators around three concepts: Trajectory predictability; Trajectory regularity; Context complexity. We compare the most common datasets used in HTP in the light of these indicators and discuss what this may imply on benchmarking of HTP algorithms. Our source code is released on Github~\footnote[1]{https://github.com/crowdbotp/OpenTraj}.
\end{abstract}


\section{Introduction}

Human trajectory prediction (HTP) is a crucial task for many applications, ranging from self-driving cars to social robots, etc. The communities of computer vision, mobile robotics, and crowd dynamics have been noticeably active on this topic. Many outstanding prediction algorithms have been proposed, from physics-based social force models~\cite{SocialForce-Helbing,ETH-Pellegrini,Yamaguchi2011} to data-driven models~\cite{SocialLSTM-Alahi,SocialGAN2018,salzmann2020}. 

In parallel, efforts have been made towards a proper benchmarking of the existing techniques. This has led to the creation of pedestrians trajectories datasets for this purpose, or to the re-use of datasets initially designed for other purposes, such as benchmarking Multiple Object Tracking algorithms. Most HTP works~\cite{Yamaguchi2011,SocialLSTM-Alahi,SocialGAN2018,SocialWays2019} report performance on the sequences of two well-known HTP datasets: the ETH dataset~\cite{ETH-Pellegrini} and the UCY dataset~\cite{Lerner2007UCY}. The metrics for comparing prediction performance involve the Average Displacement Error (ADE) and the Final Displacement Error (FDE) on standardized prediction tasks. Other datasets have been used in the same way, but performance comparisons are sometimes subject to controversy, and it remains hard to highlight how significant good performance on a particular sequence or dataset means about a prediction algorithm.  

In this paper, we address the following questions:  (1)~How to measure the complexity or difficulty of a particular dataset for the prediction task? (2)~How do the currently used HTP datasets compare to each other? Can we draw conclusions about the strengths/weaknesses of state of the art algorithms?
 
Our contributions are two-fold: (1) We propose a series of meaningful and interpretable indicators to assess the complexity behind an HTP dataset, and (2) we analyze some of the most common datasets through these indicators.

In Section~\ref{sec:notations}, we  categorize datasets complexity along three axes, trajectories predictability, trajectories regularity, and context complexity. In Section~\ref{sec:analysis}, we define indicators quantifying the complexity factors. In Section~\ref{sec:experiments}, we apply these indicators on common HTP datasets and we discuss the results in Section~\ref{sec:discussion}.

\section{Related work: HTP datasets}
Due to the non-rigidness nature of the human body or occlusions, people tracking is a difficult problem and has attracted notable attention. Many video datasets have been designed as benchmarking tools for this purpose and used intensively in HTP. Following the recent progress in autonomous driving, other datasets have emerged, involving more complex scenarios. In this section, we propose a taxonomy of HTP datasets and review some of the most representative ones.

\subsection{The zoo of HTP datasets: A brief taxonomy }
\label{subsec:zoo}
\begin{figure}[t]
	\centering
	\includegraphics[width=0.65\linewidth]{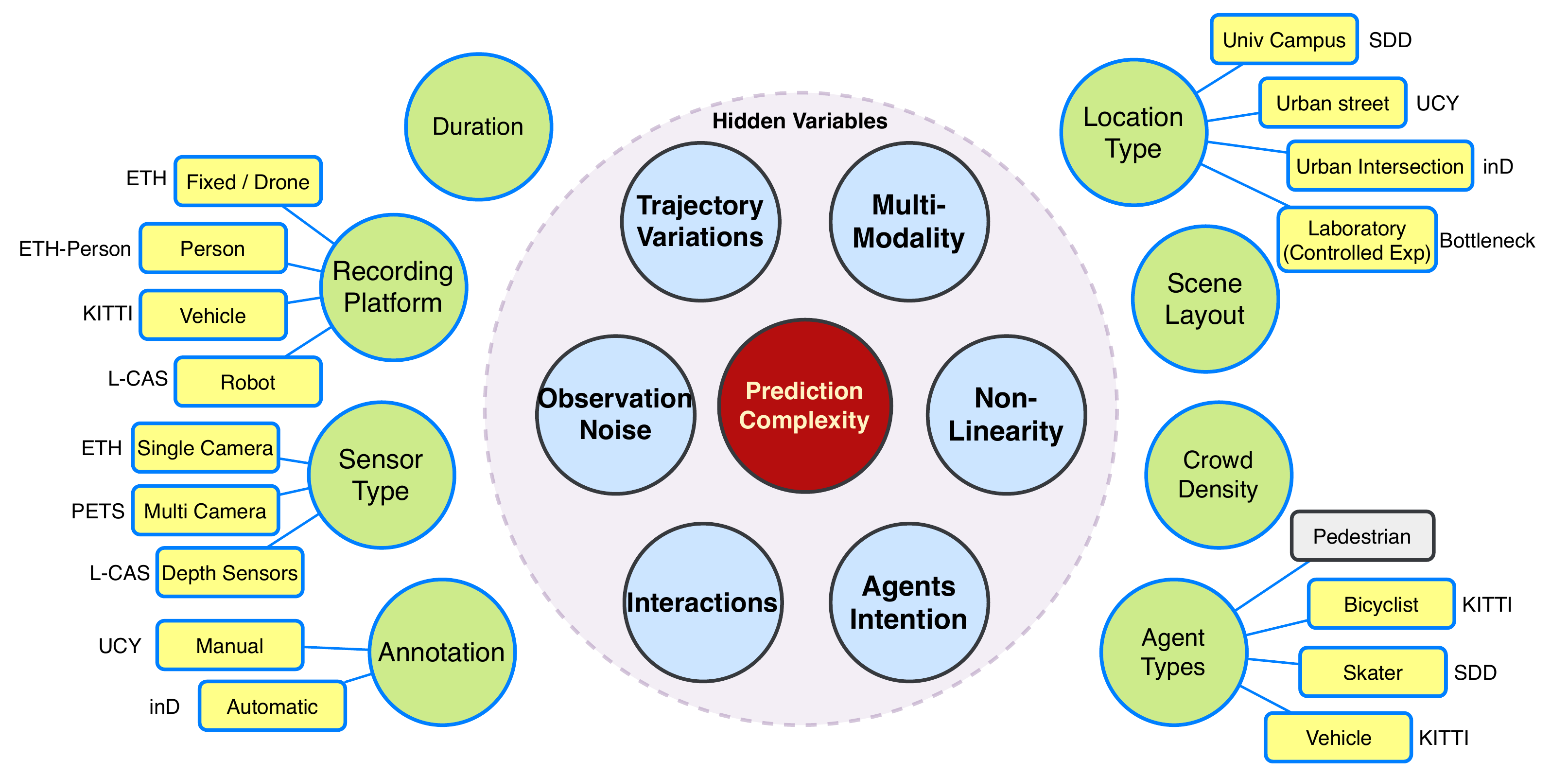}
	\caption{Taxonomy of trajectories datasets for Human Trajectory Prediction.}
	\label{fig:OpenTrajVariables}
\end{figure}

\begin{figure}[t]
	\centering	
	
	\subfloat[\scriptsize ETH-Univ]
	{\label{figur:1}\includegraphics[trim= 40 0 40 0 , clip, width=23mm] {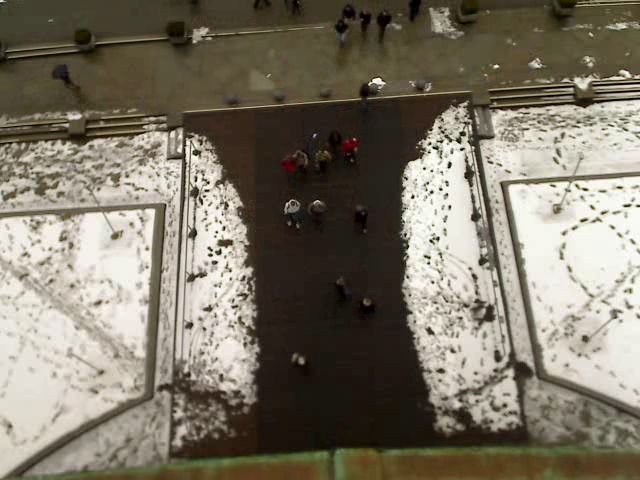}}
	\subfloat[\scriptsize ETH-Hotel]
	{\label{figur:2}\includegraphics[trim= 0 0 47 0 , clip, width=23mm] {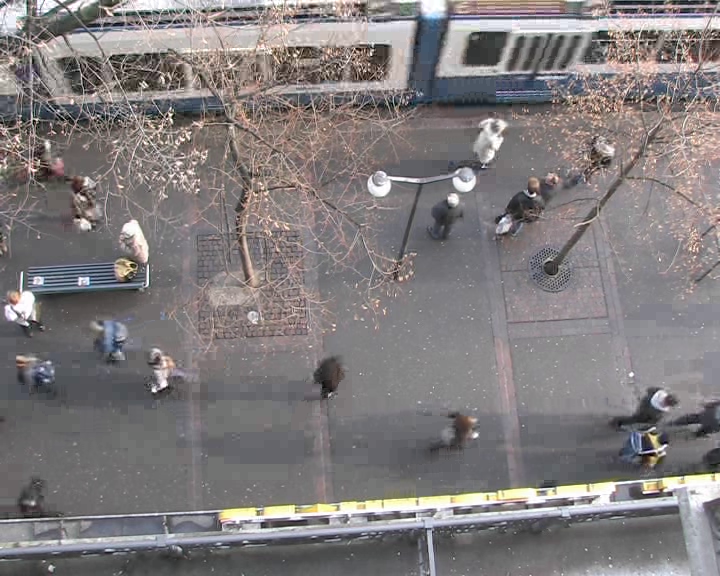}}
	\subfloat[\scriptsize UCY-Zara]
	{\label{figur:3}\includegraphics[trim= 0 0 50 0 , clip, width=23mm] {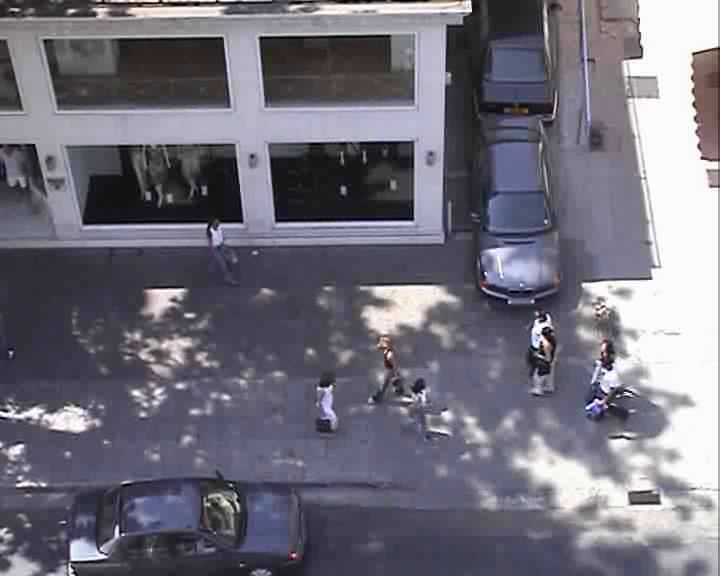}}
	\subfloat[\scriptsize UCY-Students]
	{\label{figur:4}\includegraphics[trim= 50 0 0 0 , clip, width=23mm] {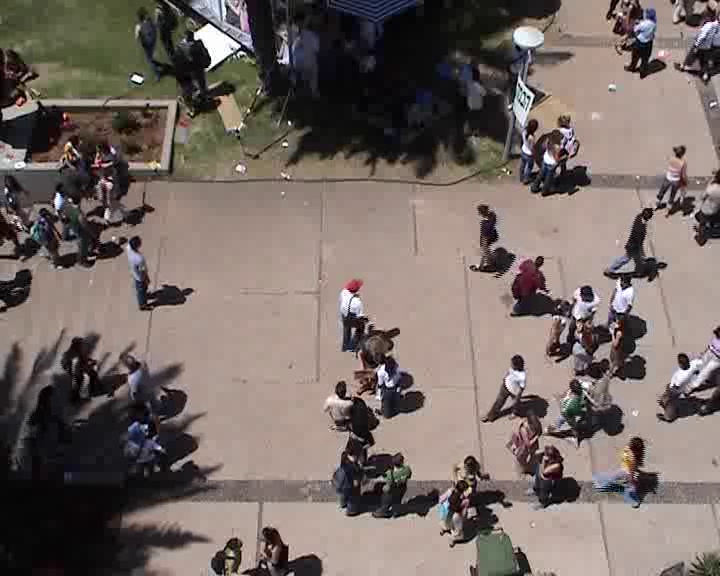}}
	\subfloat[\scriptsize UCY-Arx.]
	{\label{figur:5}\includegraphics[trim= 50 0 0 0 , clip, width=23mm] {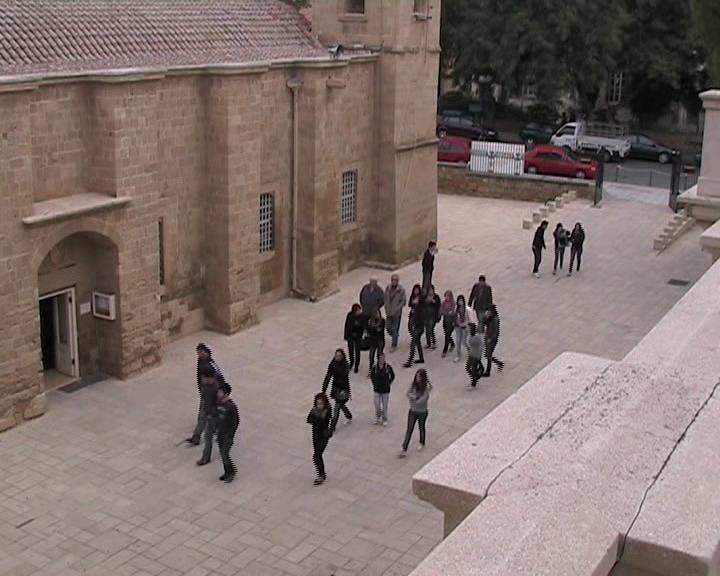}}
	\\
	\subfloat[SDD]
	{\label{figur:6}\includegraphics[trim= 320 0 0 0 , clip, width=23mm] {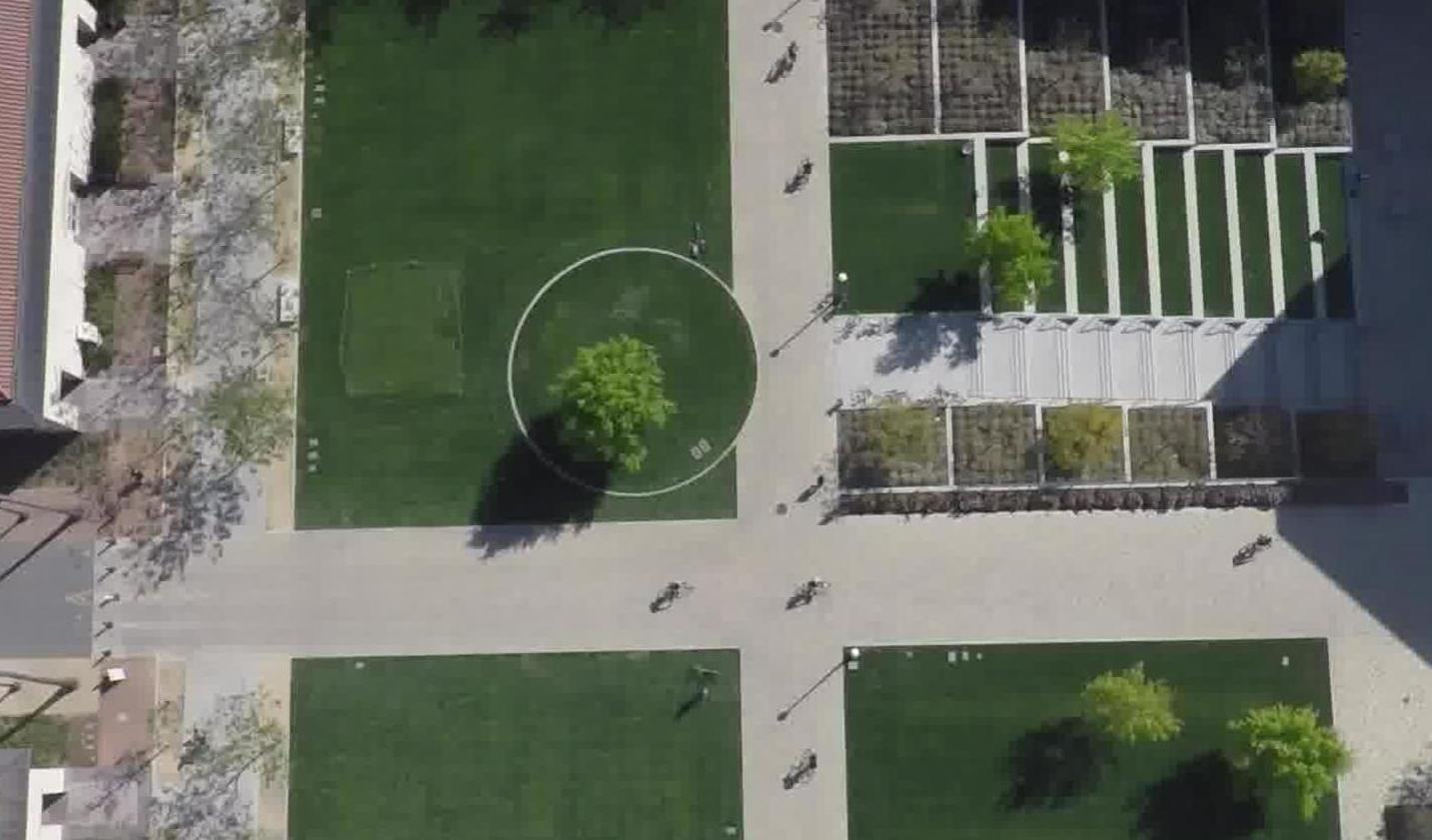}}
	\subfloat[Kitti]
	{\label{figur:7}\includegraphics[trim= 60 0 40 0 , clip, width=23mm] {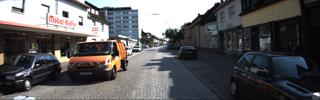}}
	\subfloat[LCas]
	{\label{figur:8}\includegraphics[trim= 60 0 40 0 , clip, width=23mm] {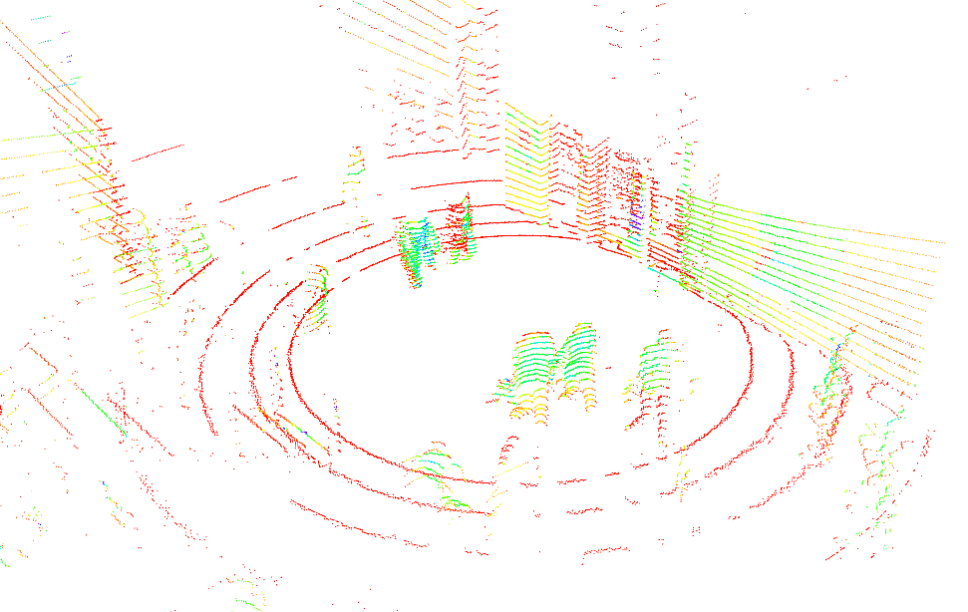}}
	\subfloat[GCS]
	{\label{figur:9}\includegraphics[trim= 300 0 170 0 , clip, width=23mm] {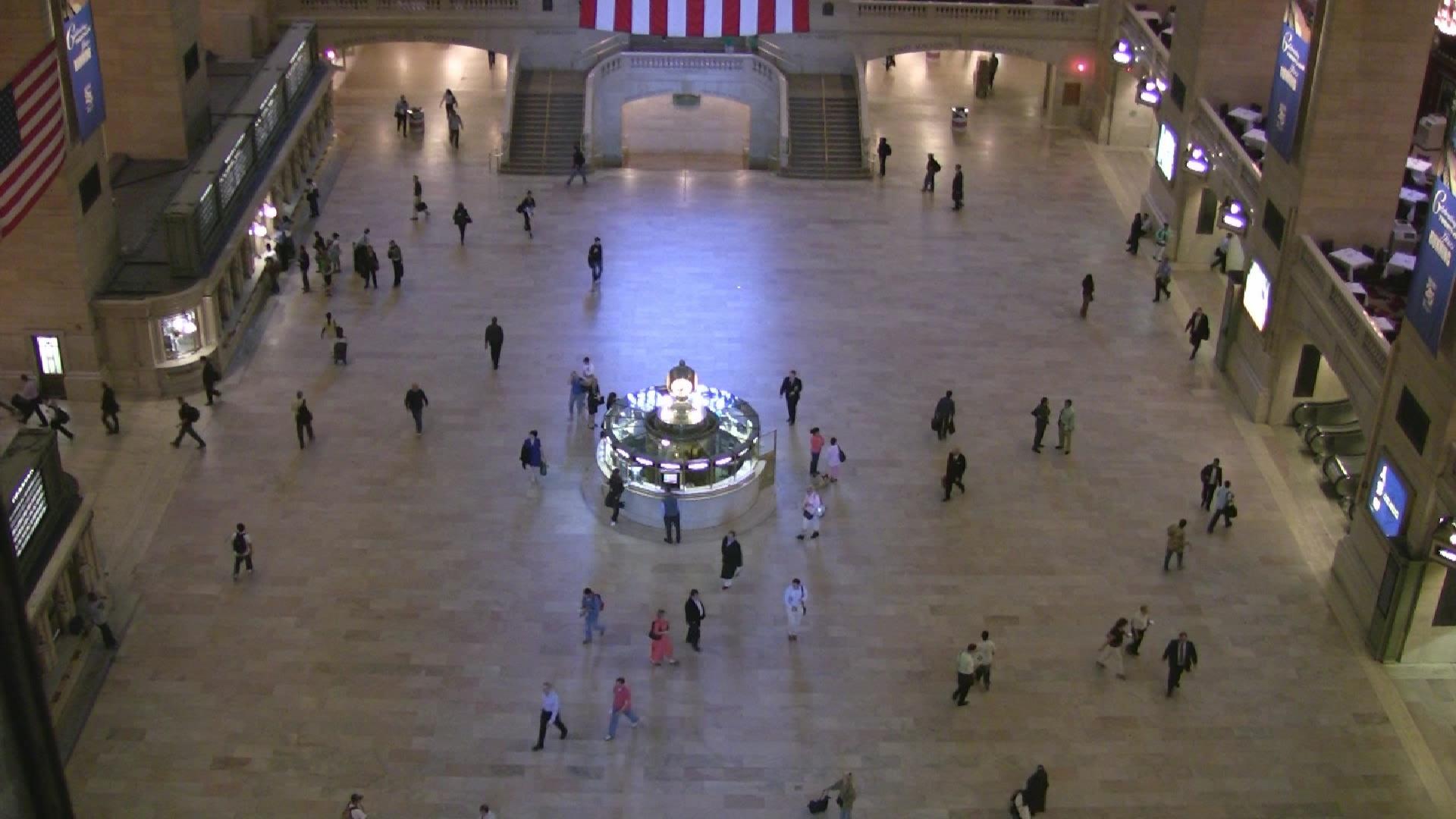}}
	\subfloat[Bottleneck]
	{\label{figur:10}\includegraphics[trim= 0 0 0 0 , clip, width=23mm] {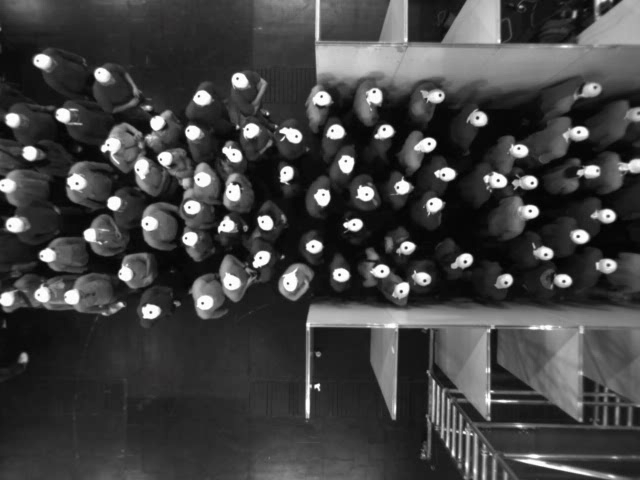}}
	
	\caption{Sample snapshots from a few common HTP datasets.}
	\label{fig:example-pics}
	
\end{figure}

Many intertwined factors explain how some trajectories or datasets are harder to predict than others for HTP algorithms. In Fig.~\ref{fig:OpenTrajVariables}, we summarize essential factors behind prediction complexity, as circles; we separate hidden (blue) and controlled (green) factors. Among hidden factors, we emphasize those related to the acquisition (noisy data), to the environment (multi-modality), or to crowd-related factors (interactions complexity). Some factors can be controlled, such as the recording platform or the choice of the location. To illustrate the variety of setups, snapshots from common HTP datasets are given in Fig.~\ref{fig:example-pics}.

Raw data may be recorded by a single~\cite{ETH-Pellegrini} or multiple~\cite{nuscenes2020} sensors, ranging from monocular cameras~\cite{inD2019,SDD-SocialEtiquette,towncentre2009} to stereo-cameras, RGB-D cameras, LiDAR, RADARs, or a mix~\cite{waymo2019,nuscenes2020}. Sensors may provide 3D annotations, but most HTP algorithms run on 2D data (the ground plane), and we focus here on 2D analysis. 

Annotation is either manual~\cite{ETH-Pellegrini,Lerner2007UCY,Kitti2013}, semi-automatic~\cite{citr2019}, or fully automatic, using detection algorithms~\cite{inD2019}. In most datasets, the annotations provide the agents' positions in the image. Annotated positions can be projected from image coordinates to world coordinates, given homographies or camera projection matrices. For moving sensors (robots~\cite{ETH-Person} or cars~\cite{Kitti2013,nuscenes2020,waymo2019}), the data are sensor-centered, but odometry data are provided to get all positions in a common frame.

\subsection{A short review of common HTP datasets}
\textbf{HTP Datasets from static cameras and drones.} The Performance Evaluation of Tracking and Surveillance (PETS) workshops have released several datasets for benchmarking Multiple Object Tracking~\cite{motchallenge2015} systems. In particular, the 11 sequences of the PETS'2009 dataset~\cite{pets2009}, recorded through 8 monocular cameras, include data from \emph{acting} pedestrians, with different levels of density, and have been used in HTP benchmarking~\cite{trajnet-sadeghian}. The Town-Centre dataset~\cite{towncentre2009} was also released for visual tracking purposes, with annotations of video footage monitoring a busy town center. It involves around two thousand walking pedestrians with  well structured (motion along a street), natural behaviors. The Wild Track dataset~\cite{Chavdarova2018} was designed for testing person detection in harsh situations (dense crowds) and provides 312 pedestrian trajectories in 400-frame sequences (from $7$ views) at 2fps. The EIF dataset~\cite{edinburgh} gives $\sim$90k trajectories of persons in a university courtyard, from an overhead camera. The BIWI pedestrian dataset~\cite{ETH-Pellegrini} is composed of 2 scenes with hundreds of trajectories of pedestrians engaged in walking activities. The ATC~\cite{Atc2013} dataset contains annotations for	92 days of pedestrian trajectories in a shopping mall, acquired from 49 3D sensors. 

The UCY dataset~\cite{UCY-CrowdsByExample} provides three scenes with walking/standing activities. Developed for crowd simulation, it exhibits different crowd density levels and a clear flow structure. The Bottleneck dataset~\cite{bottleneck-seyfried} also arose from crowd simulation and involved crowd controlled experiments (e.g., through bottlenecks). 

VIRAT~\cite{VIRAT2011} has been designed for activity recognition. It contains annotated trajectories on 11 distinct scenes, in diverse contexts (parking lot, university campus) and mostly natural behaviors. It generally involves one or two agents and objects. A particular case of activity recognition is the one of sports activities~\cite{harmon2016}, for which many data are available through players tracking technology.

The Stanford Drone Dataset (SDD)~\cite{SDD-SocialEtiquette} is a large scale dataset with 60 sequences in eight scenes, filmed from a still drone. It provides trajectories of $\sim$19k moving agents in a university campus, with interactions between pedestrians, cyclists, skateboarders, cars, buses. DUT and CITR~\cite{citr2019} datasets have also been acquired from hovering drones for evaluating inter-personal and car-pedestrian interactions. They include, respectively, 1793 and 340 pedestrian trajectories. The inD dataset~\cite{inD2019}, acquired with a static drone, contains more than 11K trajectories of road users, mostly motorized agents. The scenarios are oriented to urban mobility, with scenes at roundabouts or road intersections. Ko-PER~\cite{Koper2014} pursues a similar motivation of monitoring spaces shared between cars and non-motorized users. It provides trajectories of pedestrians and vehicles at one road intersection, acquired through laser scans and videos. Similarly, the VRU dataset~\cite{VRU2018} features around 80  cyclists trajectories, recorded at an urban intersection using cameras and LiDARs. The Forking Paths Dataset~\cite{liang2020} was created under the Carla 3D simulator, but it uses real trajectories, which are extrapolated by human annotators to simulate multi-modality with different latent goals.

\textbf{AV datasets.} Some datasets offer data collected for training/benchmarking algorithms for autonomous vehicles (AV). They may be more difficult because of the mobile data acquisition and because the trajectories are often shorter. LCAS~\cite{yz17iros} was acquired from a LiDAR sensor on a mobile robot. KITTI~\cite{Kitti2013} has been a popular benchmarking source in computer vision and robotics. Its tracking sub-dataset provides 3D annotations (cars/pedestrians) for $\sim$20 LiDAR and video sequences in urban contexts. AV companies have recently released their datasets, as Waymo~\cite{waymo2019}, with hours of high-resolution sensor data or  Argo AI with its Argoverse~\cite{Argoverse2019} dataset, featuring 3D tracking annotations for 11k tracked objects over 113 small sequences. Nutonomy disclosed its nuScenes dataset~\cite{nuscenes2020} with 85 annotated scenes in the streets of Miami and Pittsburgh.

\textbf{Benchmarking through meta-datasets.} Meta-datasets have been designed for augmenting the variety of environments and testing the generalization capacities of HTP systems. TrajNet~\cite{trajnet-sadeghian} includes ETH, UCY, SDD and PETS; in~\cite{RED}, Becker et al. proposed a comprehensive study over the TrajNet training set, giving tips for designing a good predictor and comparing traditional regression baselines vs. neural-network schemes. Trajnet++~\cite{trajnetpp2020} proposes a hierarchy of categorization among trajectories to better understand trajectory distributions within datasets. By mid-2020, over 45 solutions have been submitted on Trajnet, with advanced prediction techniques~\cite{RED,SocialLSTM-Alahi,SocialGAN2018,InteractiveGP,Giuliari2020}, but also Social-Force-based models~\cite{SocialForce-Helbing}, and variants of linear predictors, that give accuracy levels of 94\% of the best model~\cite{Giuliari2020}.
In this work, we give tools to get a deeper understanding of the intrinsic complexities behind these datasets.

\section{Problem description and formulation of needs in HTP}
\label{sec:notations}

\subsection{Notations and problem formulation}

A trajectory dataset is referred to as $\mathbb{X}$. We assume that it is made of $N_a$ trajectories of distinct agents. To be as fair as possible in our comparisons, we mainly reason in terms of absolute time-stamps, even though the acquisition frequency may vary. Within $\mathbb{X}$, the full trajectory of the $i$-th agent ($i \in [1,N_a]$) is denoted by $\textbf{T}^i$, its starting time as $\tau^i$, its duration as $\delta^i$. For $t\in [\tau^i,\tau^i+\delta^i]$, we refer to the state of agent $i$ at $t$ as $\textbf{x}^i_t$. We observe $\textbf{x}^i_t$ only for a finite subset of timestamps (at camera acquisition times). The \emph{frames} are defined as the set of observations at those times and are denoted by $\textbf{F}_t$. Each frame contains $K_t$ agents samples.

The state $\textbf{x}^i_t$ includes the 2D position $\mathbf p^i_t$ in a Cartesian system in \textit{meter}. It is often obtained from images and mapped to a world frame; the velocity $\mathbf v^i_t$, in $m / s$, can be estimated by finite differences or filtering.

To compare trajectories, following a common practice in HTP, we split all the original trajectories into $N_t$ trajlets with a common duration $\Delta=4.8s$. HTP uses trajlets of $\Delta_{obs}$ seconds as observations and the next $\Delta_{pred}$ seconds as the prediction targets. Hereafter, the set of distinct trajectories of duration $\Delta$ obtained this way are referred to as $\mathbf X^k$ where $k  \in [1,N_t]$ covers the trajlets (with potentially repetitions of the same agent). Typically, $N_t \gg N_a$. Each trajlet may be seen as an observed part and its corresponding target is referred to as $\mathbf X^k_+$.

In the following, we use functions operating at different levels, with different writing conventions. \emph{Trajectory-level functions} $F(\textbf{X})$, with capital letters, act on trajlets $\textbf{X}$. Sometimes, we consider the values of $F$ at specific time values $t$, at we denote the functions as $F_t(\textbf{X})$. \emph{Frame-level functions} $\mathcal{F}(\textbf{F})$ act on frames $\textbf{F}$.

\subsection{Datasets complexity}

We define three families of indicators over trajectory datasets that allow us to compare them and identify what makes them more ``difficult'' than other.

\textbf{Predictability.} A dataset can be analyzed through how easily individual trajectories can be predicted given the rest of the dataset, independently from the predictor. Low predictability on the trajlet distribution $p(\mathbf X)$ makes forecasting systems struggle with multi-modal predictive distributions, e.g., at crossroads. 
In that case, stochastic forecasting methods may be better than deterministic ones, as the latter typically average over the outputs seen in the training data.

\textbf{Trajectory (ir)regularity.} Another dataset characterization is through geometrical and physical properties of the trajectories, to reflect irregularities or deviations to ``simple'' models. We will use speeds, accelerations for that purpose.
	
\textbf{Context complexity.} Some indicators evaluate the complexity of the context, i.e., external factors that influence the course of individual trajectories. Typically, crowd density has a strong impact on the difficulty of HTP.  
	
These indicators operate at different levels and may be correlated. For example, complex scenes or high crowdedness levels may lead to geometric irregularities in the trajectories and to lower predictability levels. Finally, even though it is common to \emph{combine} datasets, our analysis is focused on individual datasets.

\section{Numerical Assessment of a HTP Dataset complexity}
\label{sec:analysis}

Based on the elements from Section~\ref{sec:notations}, we propose several indicators for assessing  a dataset difficulty, most of the kind $F(\mathbf X^k)$, defined at the level of trajlets $\mathbf X^k$.   
 
 \subsection{Overall description of the set of trajlets}
 \label{subsec:overall}
 \label{subsec:clusters}
To explore the distribution $p(\mathbf{T})$ in a dataset, we first consider the distributions of pedestrian positions at a timestep $t$. We parametrize each trajlet by fitting a cubic spline $\mathbf p_k(t)$ with $t \in [0, 4.8]$. For $t \in [0, 4.8]$, we get 50 time samples $\mathcal S(t) = \{ \mathbf p_k(t), \; 1 \leq k \leq N_t \}$ and analyze $\mathcal S(t)$ through clustering and entropy:
\begin{itemize}
	\item \textbf{Number of Clusters $M_t(\mathbb{X})$}: We fit a Gaussian Mixture Model (GMM) to our sample set using Expectation Maximization and select the number of clusters with the Bayesian Information Criterion~\cite{claeskens2008}. 
	
	\item \textbf{Entropy $H_t(\mathbb{X})$}: We get a kernel density estimation of $\mathcal S(t)$ (see below in Section~\ref{subsec:predictability}) and use the obtained probabilities to estimate the entropy.  
\end{itemize}

High entropy means that many data points do not occur frequently, while low entropy means that most data points are ``predictable''. Similarly, a large number of clusters would require a more complex predictive model. Both indicators give us an understanding of how homogeneous through time are all the trajectories in the dataset.
 
\subsection{Evaluating datasets trajlet-wise predictability}
\label{subsec:predictability}
To quantify the trajectory predictability, we use the conditional entropy of the predicted part of the trajectory, given its observed part. Some authors~\cite{Li2016} have used alternatively the maximum of the corresponding density. For a trajectory $\mathbf X^k \cup \mathbf X_+^k$, we define the conditional entropy conditioned to the observed $\mathbf X^k$ as

\begin{equation}
H(\mathbf{X}^k) =  -E_{\mathbf{X}_+} [\log p(\mathbf{X}_+|\mathbf{X}^k)]. 
\end{equation}

We use kernel density estimation with the whole dataset $\mathbb{X}$ ($N_t$ trajectories) to estimate it. We have $N_{obs}$ observed points during the first $\Delta_{obs}$ seconds (trajlet $\mathbf{X}_k$) and $N_{pred}$ points to predict during the last $\Delta_{pred}$ seconds (trajlet $\mathbf{X}_k^+$). We define a Gaussian kernel $K_h$ over the sum of Euclidean distances between the consecutive points along two trajectories $\mathbf{X}$ and $\mathbf{X}'$ with $N$ points each (in $\mathbb{R}^{2N}$):

\begin{equation}
K_{h,N}(\mathbf{X},\mathbf{X}') = \frac{1}{(2\pi h^2)^{N}}\exp(-\frac{1}{2h^2}\|\mathbf{X}-\mathbf{X}'\|^2),
\end{equation}
where $h$ is a common bandwidth factor for all the dimensions. We get an approximate conditional density as the ratio of the two kernel density estimates

\begin{align}
p(\mathbf{X}_+|\mathbf{X}^k) \approx
\frac{\frac{1}{N_t}\sum_{l=1}^{N_t} K_{h,N_{obs}+N_{pred}}(\mathbf{X}^k\cup\mathbf{X}_+,\mathbf{X}^l\cup\mathbf{X}^l_+)}{\frac{1}{N_t}\sum_{l=1}^{N_t} K_{h,N_{obs}}(\mathbf{X}^k,\mathbf{X}^l)}.\label{eq:predictive}
\end{align}

Since $K_{h,N_{obs}+N_{pred}}(\mathbf{X}^k\cup\mathbf{X}_+,\mathbf{X}^l\cup \mathbf{X}^l_+)=K_{h,N_{obs}}(\mathbf{X}^k,\mathbf{X}^l)K_{h,N_{pred}}(\mathbf{X}_+,\mathbf{X}_+^l)$, we can express the distribution of Eq.~\ref{eq:predictive} as the following mixture of Gaussian: 

{\scriptsize
\begin{equation}
p(\mathbf{X}_+|\mathbf{X}^k) \approx \sum_{l=1}^{N_t} \omega_l(\mathbf{X}^k) K_{h,N_{pred}}(\mathbf{X}_+,\mathbf{X}_+^l) \mbox{ with } 
 \omega_l(\mathbf{X}^k)=\frac{ K_{h,N_{obs}}(\mathbf{X}^k,\mathbf{X}^l)}{\sum_{l=1}^{N_t} K_{h,N_{obs}}(\mathbf{X}^k,\mathbf{X}^l)}.
\label{eq:condgmm}
\end{equation}
}

For a trajlet $\mathbf{X}^k$, we estimate $H(\mathbf{X}^k)$ by sampling $M$ samples $\mathbf{X}^{(m)}_+$ from Eq.~\ref{eq:condgmm}:

\begin{equation}
H(\mathbf{X}^k) \approx  -\frac{1}{M}\sum_{m=1}^{M} \log(\sum_{l=1}^{N_t} \omega_l(\mathbf{X}^k)K(\mathbf{X}^{(m)}_+,\mathbf{X}_+^l)).
\label{eq:condentropy} 
\end{equation}



\subsection{Evaluating trajectories regularity}
\label{subsec:regularity}
In this section, we define geometric and statistical indicators evaluating how \emph{regular} individual trajectories $\mathbf X^k$ in a dataset may be.

\subsubsection{Motion properties.}

A first series of indicators are obtained through \emph{speed distributions}, where speed is defined as: $s(\textbf{x}_t) = \left\Vert \mathbf v_t \right\Vert$. At the level of a trajectory $\mathbf X^k$, we evaluate the mean and the largest deviation of speeds along the trajectory
\begin{align}
S^{avg}(\mathbf X^k) = \avg_{t\in[\tau^k,\tau^k+\delta^k]}(s(\textbf{x}_t))\\ 
S^{rg}(\mathbf X^k) = \max_{t\in[\tau^k,\tau^k+\delta^k]}(s(\textbf{x}_t))- \min_{t\in[\tau^k,\tau^k+\delta^k]}(s(\textbf{x}_t)).
\label{eq:speed}
\end{align}

The higher the speed, the larger the displacements and the more uncertain the target whereabouts. Also, speed variations can reflect on high-level properties such as people activity in the environment or the complexity of this environment.

Regularity is evaluated through accelerations $a(\textbf{x}_t) \approx \frac{1} {dt} [s(\textbf{x}_{t+dt}) - s(\textbf{x}_{t})]$. It can reflect the interactions of an agent with its environment according to social-force model~\cite{SocialForce-Helbing}: agents typically keep their preferred speed while there is no reason to change it. High accelerations appear when an agent avoids collision or joins a group. We consider the average and maximal accelerations along $\mathbf X^k$

\begin{equation}
A^{avg}(\mathbf X^k) = \avg_{t\in[\tau^k,\tau^k+\delta^k]}(|a(\textbf{x}_t)|);\;\;
A^{max}(\mathbf X^k) = \max_{t\in[\tau^k,\tau^k+\delta^k]}(|a(\textbf{x}_t)|).
\label{eq:acc}
\end{equation}		

\subsubsection{Non-linearity of trajectories.} \emph{Path efficiency} is defined as the ratio of the distance between the trajectory endpoints over the trajectory length:
\begin{equation}
F(\textbf{X}^k) = \frac{\left\Vert p_{\tau^k + \delta^k} - p_{\tau^k} \right\Vert}
{\int_{t=\tau^k}^{\tau^k + \delta^k} dl}.
\label{eq:eff}
\end{equation}	
The higher its value, the closer the path is to a straight line, so we would expect that the prediction task will be ``easier'' for high values of $F(\textbf{X}^k)$. 

Another indicator is the average angular deviation from a linear motion. To estimate it, we align all trajlets by translating them to the origin of coordinate system and rotating them such that the first velocity is aligned with the $x$ axis:
\begin{align}
\mathbf{\hat X}^k = 
\begin{bmatrix}
\mathbf R({-\measuredangle \mathbf{v}_0^k}) & -\mathbf{p}_0^k 
\end{bmatrix} 
\begin{bmatrix}
\mathbf{X}^k \\ 1
\end{bmatrix} ^ T.
\end{align}
Then the deviation of a trajectory $\mathbf X^k$ at $t$ and its average value are  defined as:

\begin{equation}
D_t(\mathbf X^k) =  {\measuredangle\mathbf{\hat X}_t^k} \mbox{  and  } D(\mathbf X^k) =  \avg_{t\in[\tau^k,\tau^k+\delta^k]}(D_t(\mathbf X^k)).
\label{eq:dev}
\end{equation}

\subsection{Evaluating the context complexity}
\label{subsec:context}
The data acquisition context may impact HTP in different ways. It may ease the prediction by introducing correlations: With groups, it can be easier to predict one's motion from the other group members. In general, social interactions result into adjustments that may be generate non-linearities (and lower predictability). 

\subsubsection{Collision avoidance} is the most basic type of interaction. Higher density resulting into more interactions, this aspect is also evaluated by the density metrics below. However, high-density crowds may ease the prediction (e.g., laminar flow of people). To reflect the intensity of collision avoidance-based interactions, we use the \emph{distance of closest approach} (DCA)~\cite{olivier2013} at $t$, for a pair of agents $(i,j)$:

\begin{equation}
\dca(t,i,j)=\sqrt{\| \mathbf{x}^i_t-\mathbf{x}^j_t \|^2 - (\max(0,\frac{( \mathbf{v}^i_t-\mathbf{v}^j_t)^T( \mathbf{x}^i_t-\mathbf{x}^j_t)}{\| \mathbf{v}^i_t-\mathbf{v}^j_t \|}))^2},
\end{equation}
and for a trajlet $\mathbf X^k$ (relative to an agent $i_k$), we consider the overall minimum

\begin{equation}
C(\mathbf X^k) = \min_{t\in[\tau^k,\tau^k+\delta^k]} \min_j \dca(t,i_k,j).
\label{eq:dca}
\end{equation} 

In~\cite{Karamouzas2014}, the authors suggest that time-to-collision (TTC) is strongly correlated with trajectory adjustments. The TTC for a pair of agents $i,j$, modeled as disks of radius $R$, for which a collision will occur when keeping their velocity, is 

\begin{equation}
\tau(t,i,j)=\frac{1}{\| \mathbf{v}^i_t-\mathbf{v}^j_t \|^2}[\delta^{ij}_t-\sqrt{(\delta^{ij}_t)^2-\| \mathbf{v}^i_t-\mathbf{v}^j_t \|^2(\| \mathbf{x}^i_t-\mathbf{x}^j_t \|^2-4R^2)}]
\label{eq:ttc}
\end{equation}
where $\delta^{ij}_t=( \mathbf{v}^i_t-\mathbf{v}^j_t)^T( \mathbf{x}^i_t-\mathbf{x}^j_t)$. In~\cite{Karamouzas2014}, the authors also proposed quantifying the interaction strength between pedestrians as an energy function of $\tau$:

\begin{equation}
E({\tau}) = \frac{k}{\tau^2}e^{-\frac{\tau}{\tau^+}},
\label{eq:energy}
\end{equation}
with $k$ a scaling factor and $\tau^+$ an upper bound for TTC. Like~\cite{Karamouzas2014}, we estimate the actual TTC probability density  between pedestrians (from Eq.~\ref{eq:ttc}) over the probability density that would arise without interaction (using the time-scrambling approach of~\cite{Karamouzas2014}). Then we estimate $E(\tau)$ with Eq.~\ref{eq:energy}. As the range of well-defined values for $\tau$ may be small, we group the data into $0.2s$ intervals and use t-tests to find out the lower bound $\tau^-$ when two consecutive bins are significantly different $(p<0.05)$. The upper bound $\tau^+$ is fixed as $3s$. TTC and energy interaction are extended for trajlets (only if there exists future collision):

\begin{equation}
T(\mathbf X^k) = \min_{t\in[\tau^k,\tau^k+\delta^k]} \min_j \tau(t,i_k,j) \mbox{ and } E(\mathbf X^k)=E(T(\mathbf X^k)) .
\label{eq:ttctrajlet}
\end{equation}

\subsubsection{Density \& distance measures.} For a frame $\textbf{F}_t$, the \emph{Global Density} is defined as the number of agents per unit area $\mathcal D(\textbf{F}_t) = \frac{K_t}{\textbf{A}(\mathbb X)}$,
with $K_t$ the number of agents present at $t$ and $\textbf{A}(\mathbb X)$ the spatial extent of $\mathbb X$, evaluated from the extreme $x, y$ values. The \emph{Local Density} measures the density in a neighborhood. Plaue et al.~\cite{Plaue2011} infer it with a nearest-neighbour kernel estimator. For a point $\mathbf{x}_t$,
 
\begin{equation}
\rho(\mathbf{x}_t) = \frac{1}{2\pi}\sum_{i=1}^{K_t}\frac{1}{(\lambda d_t^i)^2}\exp{\left(-\frac{\|{\mathbf{x}^i_t-\mathbf{x}_t}\|^2}{2(\lambda d_t^i)^2}\right)},
\end{equation}
with $d_t^i=\min_{j\neq i} \|{\mathbf{x}^i_t-\mathbf{x}^j_t}\|$ the  distance from $i$ to its nearest neighbor and $\lambda>0$ a smoothing parameter. $\rho$ is used to evaluate a trajlet-wise local density indicator

\begin{equation}
L(\mathbf X^k) =  \max_{t\in[\tau^k,\tau^k+\delta^k]} \rho(\mathbf{x}^{i_k}_t).
\label{eq:localdensity}
\end{equation}
		

%
%

\section{Experiments}
\label{sec:experiments}
{\scriptsize
	\begin{table}[t] 
		\centering
		\caption{General statistics of assessed datasets. The columns present the type of location where the data is collected, the acquisition means, number of annotated pedestrians, the -rounded- duration (in \underline{m}inute or \underline{h}our), the total duration of all trajectories, number of trajlets, and percent of non-static trajlets, respectively.	}
		\begin{tabular}{ c p{2.5cm} | c c| p{0.6cm}p{0.8cm}p{0.7cm}| cc}

			\multicolumn{2}{c}{Dataset}
			& Location
			& \multicolumn{1}{c}{Acquisition}
			
			& \mcrot{1}{l}{35}{\#peds} 	
			& \mcrot{1}{l}{35}{duration} 	
			& \mcrot{1}{l}{35}{total dur.}	
			
			& \mcrot{1}{l}{35}{\#trajlets}
			& \mcrot{1}{l}{35}{non-static} 
			\\	\midrule \midrule
			
			\multirow{2}{*}{\rotatebox{90}{\textbf{ETH}}}
			& Univ 
			& univ entrance &  \multirow{2}{*}{top-view cam}  
			& 360 & 13m & 1h 
			& 823 & 93\% \\
			
			& Hotel 
			& urban street &    
			& 390 & 12m & 0.7h
			& 484 & 66\% \\
			\midrule
			
			\multirow{2}{*}{\rotatebox{90}{\textbf{UCY}}}
			& Zara 
			& urban street & \multirow{2}{*}{top-view cam}
			& 489 & 18m & 2.1h 
			& 2130 & 75\% \\
			
			& Students 
			& univ campus &    
			& 967 & 11.5m & 4.5h 
			& 4702 & 96\% \\
			\midrule		
			
			\multirow{3}{*}{\rotatebox{90}{\textbf{SDD}}}
			
			& Coupa 
			& \multirow{3}{*}{univ campus} & \multirow{3}{*}{drone cam}
			& 297 & 26m & 4.5h
			& 5,394 & 41\% \\
			
			& Bookstore 
			& & 	
			& 896 & 56m & 9.5h 
			& 11,239 & 54\% \\		
			
			& DeathCircle
			& &
			& 917 & 22.3m & 4.2h
			& 8,288 & 62\% \\
			
			\midrule		
			
			\multirow{2}{*}{\rotatebox{90}{\textbf{inD}}}
			& inD-Loc(1) 
			& \multirow{2}{*}{urban intersection } & \multirow{2}{*}{drone cam}
			& 800 & 180m & 7.1h
			& 8302 & 94\% \\
			
			& inD-Loc(2)
			& &
			& 2.1\textbf{k} & 240m & 18h 
			& 21234 & 95\% \\
			\midrule

			\multirow{3}{*}{\rotatebox{90}{\tiny{Bottleneck}}}
			& 1D Flow\tiny{(w=180)}
			& \multirow{2}{*}{\small{simulated corridor}} & \multirow{2}{*}{top-view cam}
			& 170 & 1.3m & 1h
			& 940 & 99\% \\
			
			& 2D Flow\tiny{(w=160)}
			& &
			& 309 & 1.3m & 1.5h
			& 1552 & 100\% \\
			\midrule

			~
			& \textbf{Edinburgh} Sep\{1,2,4,5,6,10\}
			& \multirow{2}{*}{univ forum} & \multirow{2}{*}{top-view cam}
			& \multirow{2}{*}{1.2\textbf{k}} & \multirow{2}{*}{9h} & \multirow{2}{*}{3h}
			& \multirow{2}{*}{2124} & \multirow{2}{*}{83\%} \\		
			\midrule
			
			~
			& \textbf{GC Station}
			& train station & surveillance cam      
			& 17\textbf{k} & 1.1h & 79h
			& 76866 & 99\% \\
			\midrule
			
			
			\multirow{1}{*}{\rotatebox{90}{\textbf{~}}}
			& \textbf{Wild-Track}
			& univ campus & multi-cam 
			& 312 & 3.3m & 1.3h
			& 1215 & 57\% \\
			\midrule
			
			\multirow{2}{*}{\rotatebox{90}{\textbf{~}}}
			& \textbf{KITTI}
			& urban streets & lidar\& multi-cam
			& 142 & 5.8m & 0.3h
			& 253 & 93\% \\
			\midrule

			\multirow{1}{*}{\rotatebox{90}{\textbf{~}}}
			& \textbf{LCas}-Minerva
			& univ-indoor & lidar
			& 878 & 11m & 4.8h
			& 3553 & 83\% \\
			\midrule
		\end{tabular}
		\label{tab:datasets}
	\end{table}
}

In this section, we analyze some common HTP datasets in the light of the indicators presented in the previous section. In Table~\ref{tab:datasets}, we give statistics (location, number of agents, duration\dots) for the datasets we have chosen to evaluate. We gather the most commonly used in HTP evaluation (ETH, UCY, SDD in particular) and datasets coming from a variety of modalities (static cameras, drones, autonomous vehicles\dots), to include different species from the zoo of Section~\ref{subsec:zoo}. 

For those including very distinct sub-sequences, e.g., ETH, UCY, SDD, inD, and Bottleneck (also denoted by BN in the figures), we split them into their constituting sequences. Also, note that we have focused only on pedestrians (no cyclist nor cars). We also ruled out any dataset containing less than 100 trajectories (e.g., UCY Arxiepiskopi or PETS).

To analyze a dataset $\mathbb{X}$, we apply systematically the following preprocessing
\begin{enumerate}
\item Projection to world coordinates, when necessary. 
\item Down-sampling the annotations to a 2-3 fps framerate; 
\item Application of a Kalman smoothing with a constant acceleration model; 
\item Splitting of the resulting trajectories into trajlets $\mathbf X^k$ of length $\Delta=4.8$s and filtering out trajlets shorter than $1$m.
\end{enumerate}
We finally recall the trajlet-wise indicators we have previously introduced:

\begin{tabular}{|p{3cm}|p{8cm}|}
		\hline Overall description & Entropy $H_t(\mathbb{X}^k)$ and clusters $M_t(\mathbb{X})$ (section~\ref{subsec:clusters}).\\		
		\hline Predictability & Cond. entropy $H(\mathbf{X}^k)$ (Eq.~\ref{eq:condentropy}).\\
		\hline Regularity & Speed $S^{avg}(\mathbf X^k),S^{rg}(\mathbf X^k)$ (Eq.~\ref{eq:speed}).\\
		& Acceleration $A^{avg}(\mathbf X^k),A^{max}(\mathbf X^k)$ (Eq.~\ref{eq:acc}).\\
		& Efficiency $F(\textbf{X}^k)$ (Eq.~\ref{eq:eff}).\\
		& Angular deviation $D(\textbf{X}^k)$ (Eq.~\ref{eq:dev}).\\
		\hline Context & Closest approach $C(\mathbf X^k)$ (Eq.~\ref{eq:dca}).\\
		& Time-to-collision $T(\mathbf X^k)$, energy $E(\mathbf X^k)$ (Eq.~\ref{eq:ttctrajlet}).\\
		& Local density $L(\mathbf X^k)$ (Eq.~\ref{eq:localdensity}).\\
		\hline
\end{tabular}

\begin{figure}[t]
\centering
\includegraphics[width=0.9\linewidth]{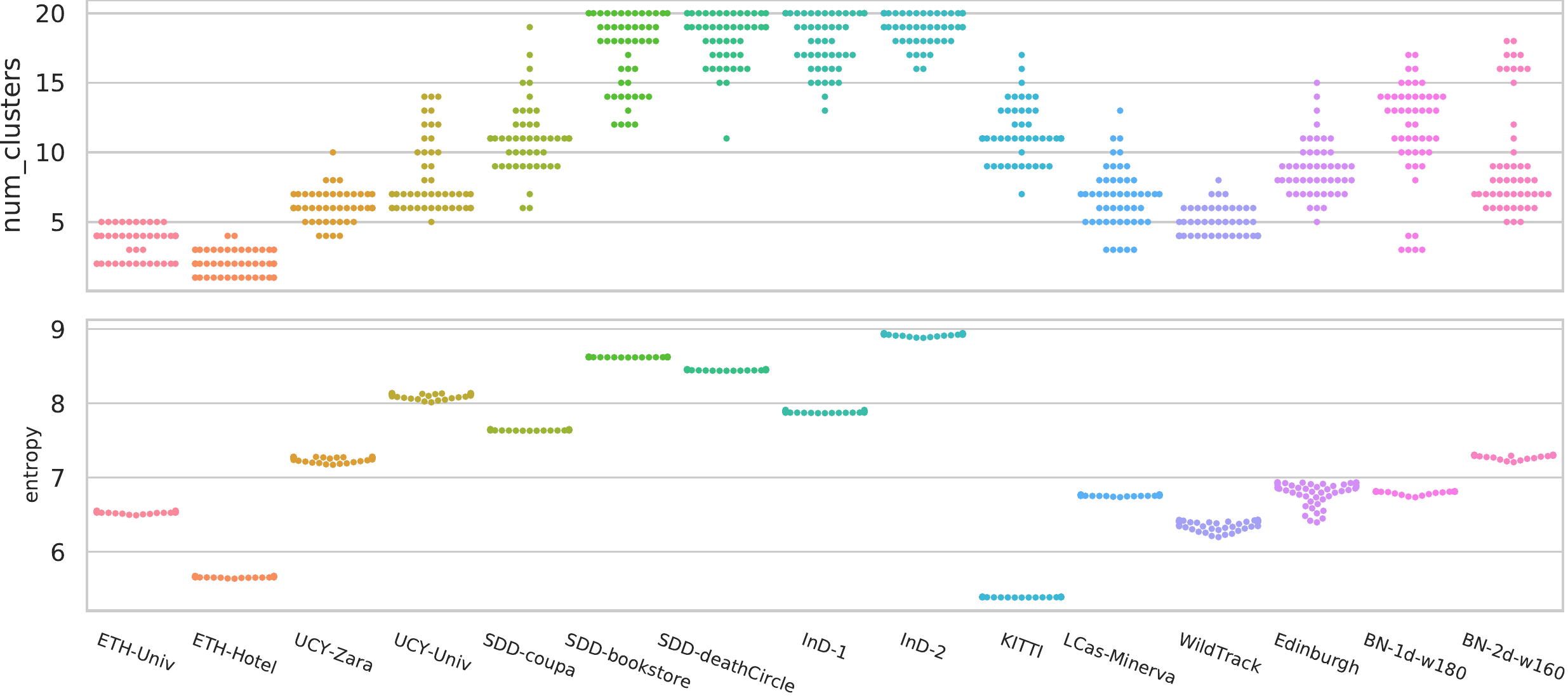}
\caption{Entropy $H_t(\mathbb{X})$ and number of clusters $M_t(\mathbb{X})$, as described in Section~\ref{subsec:clusters}, at different progression rates $t$, for a dataset $\mathbb{X}$. Each dot corresponds to one $t$\label{fig:distribution-entropyclusters}.}
\end{figure}

\paragraph{Overall description of the set of trajlets.} For the indicators of Section~\ref{subsec:predictability}, we have chosen $h=0.5m$ for the Gaussian in the kernel-based density estimation; the number of samples used to evaluate the entropy is $M=30$; the maximal number of clusters when clustering unconditional or conditional trajectories distributions is $21$. In Fig.~\ref{fig:distribution-entropyclusters}, we plot the distributions of the overall entropy and number of clusters, at different progression rates along the dataset trajectories. Without surprise, higher entropy values are observed for the less structured datasets (without main directed flows) such as SDD or inD. The number of clusters follows a similar trend, indicating possible multi-modality.
\begin{figure}
	\centering
	\includegraphics[trim= 0 0 0 143 , clip, width=1.02\linewidth]{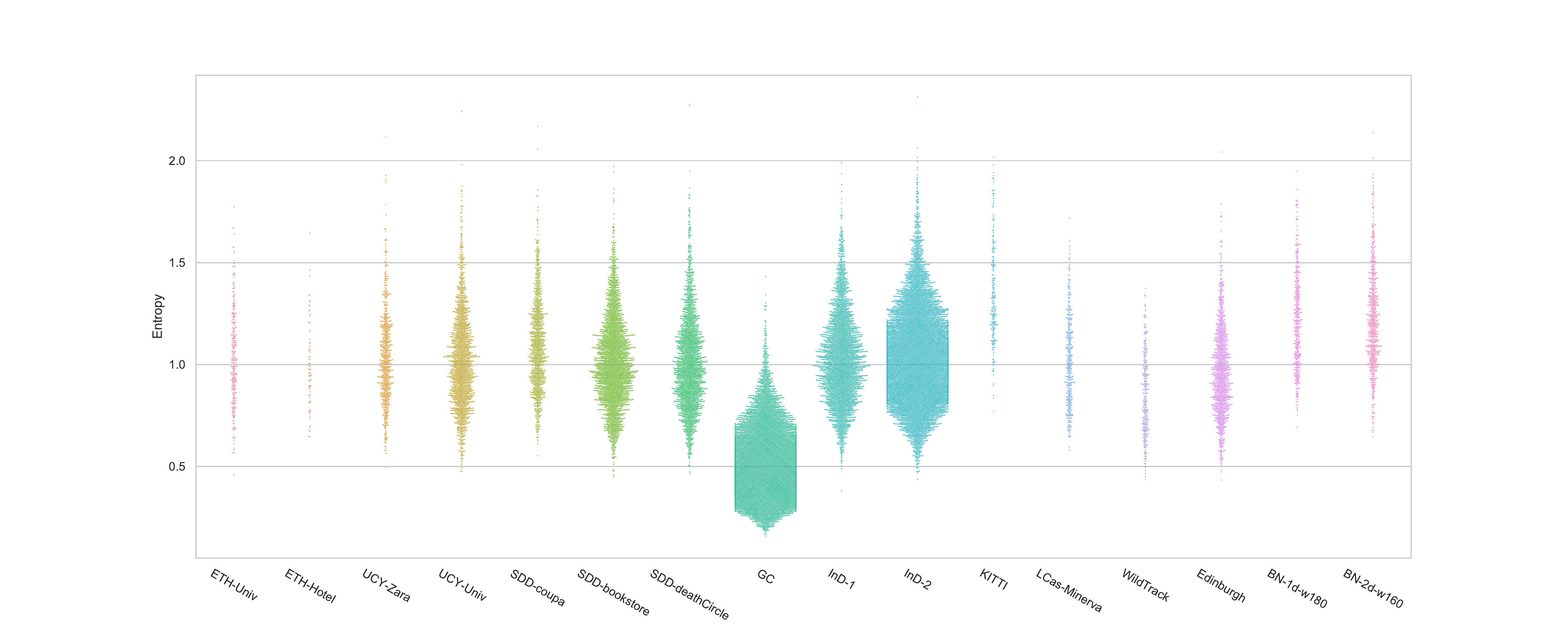}
	\caption{Conditional Entropies $H(\mathbf{X}^k)$.}
	\label{fig:Conditional-entropies}
\end{figure}

\paragraph{Predictability indicators.} In 
Fig.~\ref{fig:Conditional-entropies}, we depict the values of $H(\mathbf{X}^k)$, with one dot per trajlet $\mathbf{X}^k$. Interestingly, excepting the Bottleneck sequences, where high density generates randomness, the support for the entropy distributions are similar among datasets. What probably makes the difference are the tails in these distributions: large lower tails indicate high proportions of easy-to-predict trajlets, while large upper tails indicate high proportions of hard-to-predict trajlets.

\begin{figure}[t]
\centering
\includegraphics[width=0.9\linewidth]{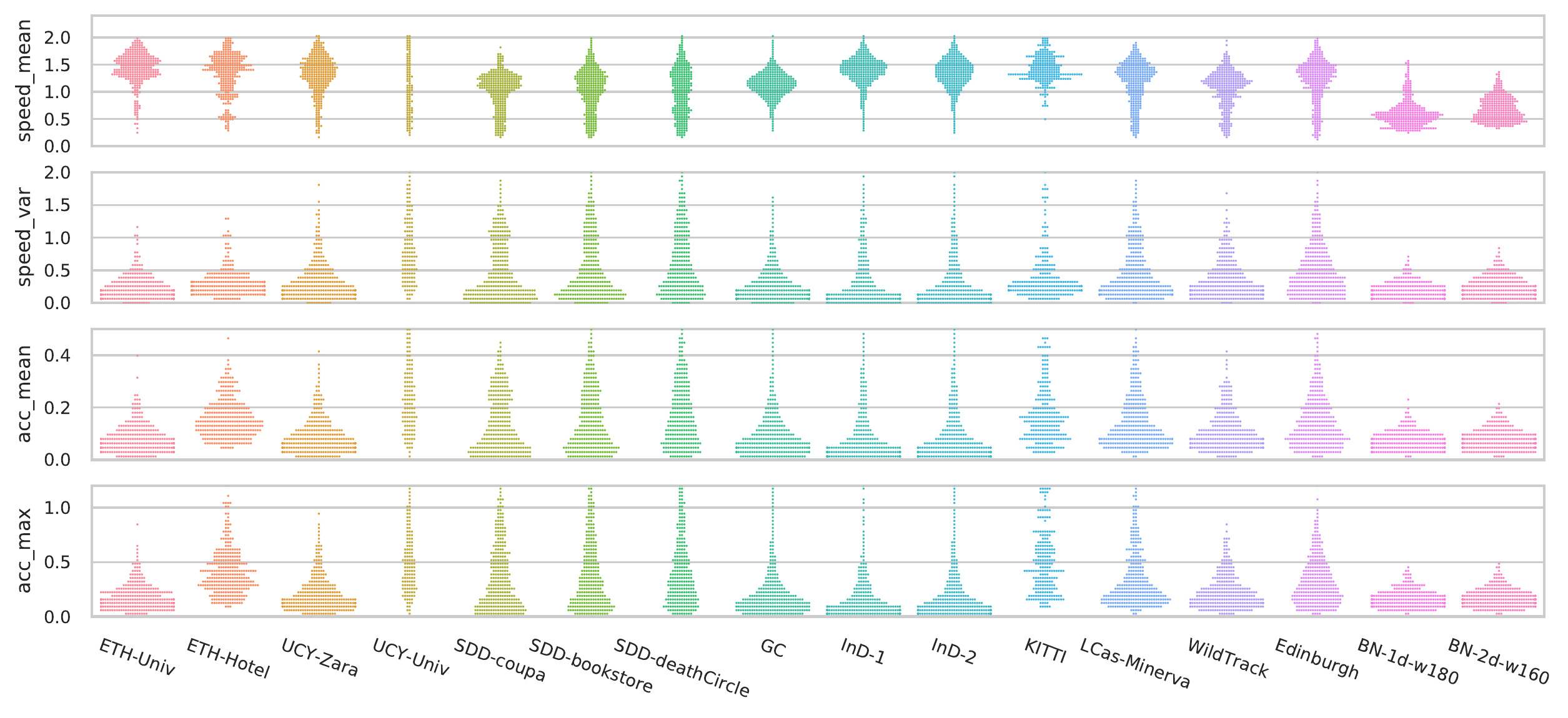}
\caption{Speed and acceleration indicators $S^{avg}(\mathbf X^k),S^{rg}(\mathbf X^k),A^{avg}(\mathbf X^k),A^{max}(\mathbf X^k)$. From top to bottom: speed means and variations, mean and max. accelerations.}
\label{fig:motion_metric}
\end{figure}

\begin{figure}
\subfloat[Path Efficiency. The higher, the closer to a straight line.]{
\centerline{\includegraphics[width=0.9\linewidth]{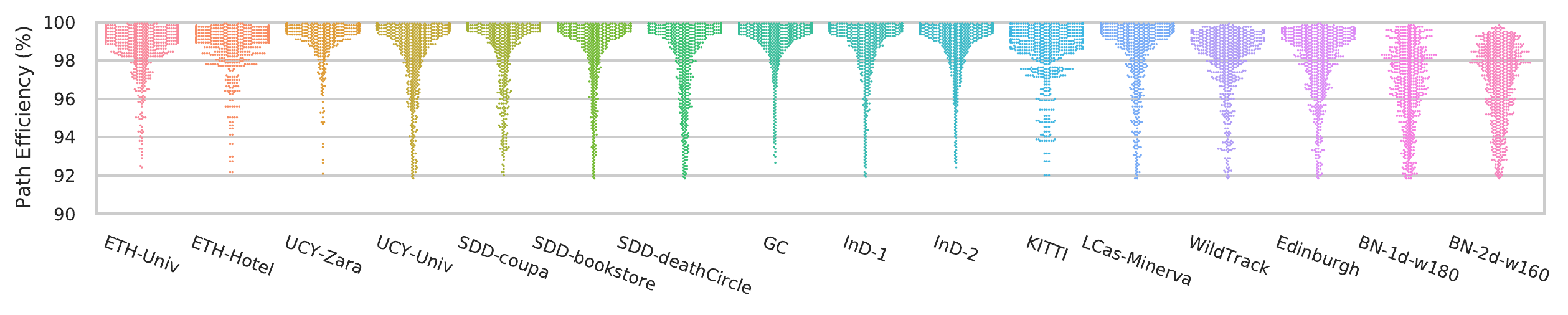}}
\label{fig:path_eff}}

\subfloat[Deviation from linear motion]{
	\centerline{\includegraphics[width=0.8\linewidth]{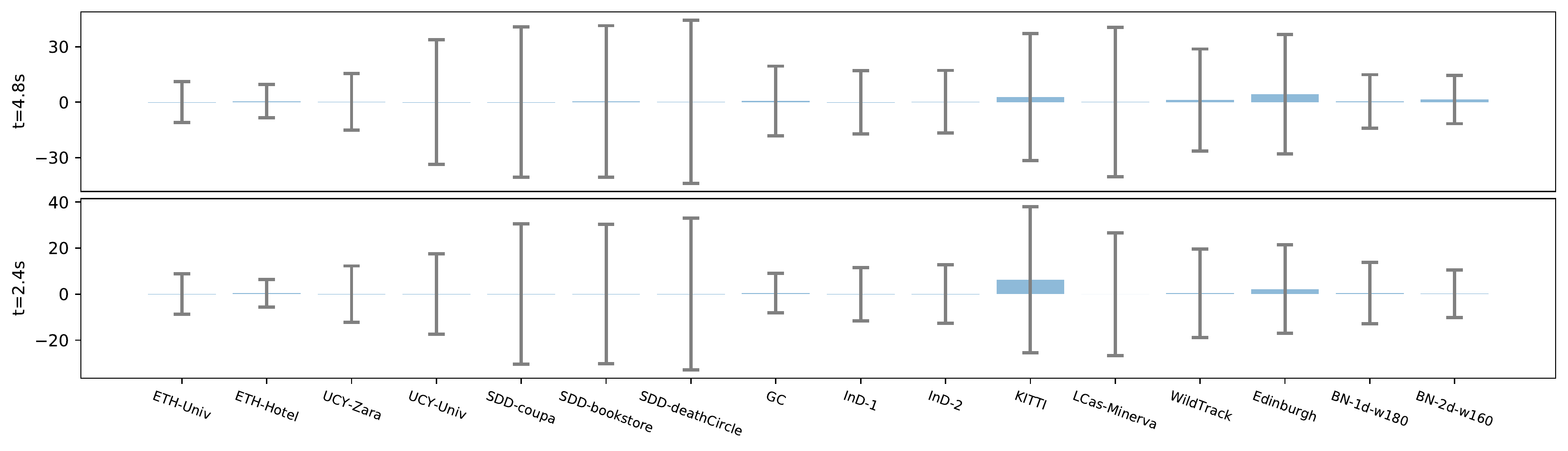}}
	\label{fig:deviation}
}	
\caption{Regularity indicators: Path efficiency and deviation from linear motion.}
\end{figure}

\paragraph{Regularity indicators.} In Fig.~\ref{fig:motion_metric}, we depict the distributions of the regularity indicators $S^{avg}(\mathbf X^k),S^{rg}(\mathbf X^k),A^{avg}(\mathbf X^k),A^{max}(\mathbf X^k)$ from Eqs.~\ref{eq:speed} and~\ref{eq:acc}. Speed averages are generally centered around $1$ and $1.5m/s$. Disparities among datasets appear with speed variations and average accelerations: ETH or UCY Zara sequences do not exhibit large speed variations, e.g. compared to Wild Track. In Fig.~\ref{fig:path_eff}, we depict the path efficiency $F(\textbf{X}^k)$ fro Eq.~\ref{eq:eff}, and we observe that ETH, UCY paths tend to be straighter. More complex paths appear in Bottleneck, due to the interactions within the crowd, or in SDD-deathCircle, EIF, due to the environment complexity. In Fig.~\ref{fig:deviation}, deviations $D_t(\mathbf X^k)$ are displayed for different progression rates along the trajectories, and reflect similar trends.

\begin{figure}
	\centerline{\includegraphics[width=0.9\linewidth]{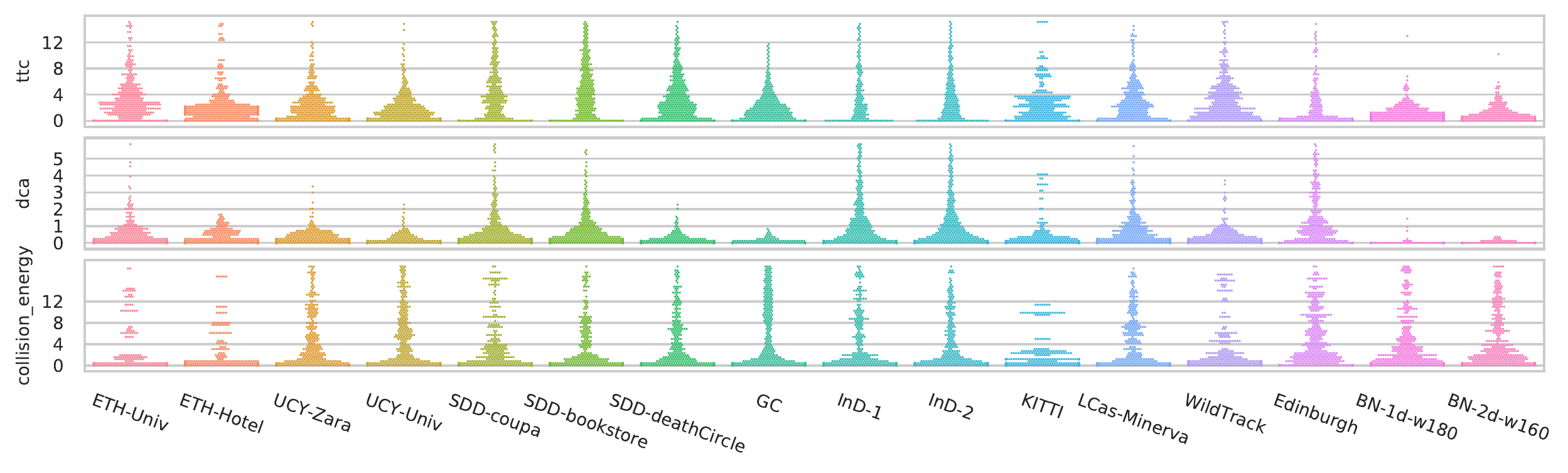}}
	\caption{Collision avoidance-related indicators:  From top to bottom, time-to-collision, Distance of closest approach and Interaction energy. \label{fig:distribution-collision}}
\end{figure}	

\begin{figure}
	\centerline{\includegraphics[width=0.9\linewidth]{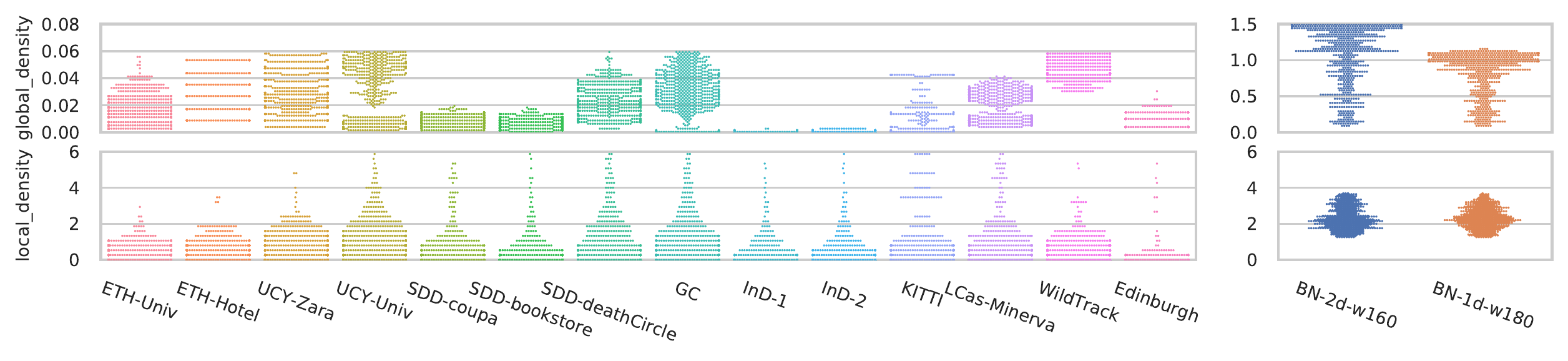}}
	\caption{Density indicators: On the top, global density (one data point for each frame); on the bottom, local density (one data point for each trajlet). }
	\label{fig:distribution-density}
\end{figure}	

\paragraph{Context complexity indicators.} For estimating the TTC in Eq.~\ref{eq:ttc}, we set $R=0.3m$, and for the interaction energy of Eq.~\ref{eq:energy}, we set $k=1$. The local density of Eq.~\ref{eq:localdensity} uses $\lambda=1$. In Fig.~\ref{fig:distribution-collision}, we display the collision avoidance-related indicators (TTC, DCA and interaction energy) described in Section~\ref{subsec:context}, while in Fig.~\ref{fig:distribution-density}, we depict the density-related indicators. Most samples have low interaction energy, but interesting interaction levels are visible Zara, InD. The global density for most datasets stays less than 0.1 $p/m^2$ while in InD(1\&2), Edinburgh and SDD (Coupa \& Bookstore), it is even less than $0.02$. Bottleneck (1d 
\& 2d) are significantly high density scenarios. For this reason why we depict them separately. Most natural trajectory datasets have a local density about $0-4 p/m^2$ while such number is higher ($2-4 p/m^2$) in Bottleneck. With both density indicators, a dataset such as WildTrack has a high global density and low local density, indicating a relatively sparse occupation. Conversely, low global density and high local density in Ind suggests the pedestrians are more clustered. This observation is also reflected in the interaction and entropy indicators as well.

\section{Discussion}
\label{sec:discussion}
Among the findings from the previous Section, Fig.~\ref{fig:Conditional-entropies} shows that the predictability among most datasets varies in mostly the same ranges.
Regarding the motion properties of the datasets (see Fig.~\ref{fig:motion_metric}), another finding is pedestrians' average speed, which, in most cases, varies from 1.0 to 1.5 m/s. However, this is not the case for \textit{Bottleneck} dataset, because the high density of the crowd does not allow the pedestrians to move with a `normal` speed. In the SDD dataset, we observe multiple pedestrians strolling the campus. As shown in Fig.~\ref{fig:deviation} these low-speed motions are usually associated with high deviation from linear motion, though part of this effect is related to the complexity of the scene layout.

Also, for most of the datasets, the speed variation of trajlets remains almost below 0.5. This is not a true hypothesis for \textit{LCas} and \textit{WildTrack}. As one would expect, the distribution of mean/max acceleration of trajlets is highly correlated with speed variations. In Fig.~\ref{fig:path_eff} we see that almost all values are bigger than 90\%. For \textit{Bottleneck} we see this phenomenon, where by increasing the crowd density and decreasing crowd speed, the paths become less efficient.

\section{Conclusions \& Future Work}

We have presented in this work a series of indicators for gaining insight into the intrinsic complexity of Human Trajectory Prediction datasets. These indicators cover concepts such as trajectory predictability and regularity, and complexity in the level of inter-pedestrian interactions. In light of these indicators, datasets commonly used in HTP exhibit very different characteristics. In particular, it may explain why predictions techniques that do not use explicit modeling of social interactions, and consider trajectories as independent processes, may be rather successful on datasets where e.g., most trajectories have low collision energy; it may also indicate that some of the more recent datasets with higher levels of density and interaction between agents could provide more reliable information on the quality of the prediction algorithm. Finally, the trajlet-wise analysis presented here opens the door to some evolution in benchmarking processes, as we could evaluate scores by re-weighting the target trajlets in the function of the presented indicators.   

\section*{Acknowledgements}   

This research is supported by the CrowdBot H2020 EU Project
\url{http://crowdbot.eu/} and by the Intel Probabilistic Computing initiative. The work done by Francisco Valente Castro was sponsored using an MSc Scholarship given by CONACYT with the following scholar registry number 1000188.


%
%
\bibliographystyle{splncs}
\bibliography{refs}
\end{document}